\documentclass[letterpaper]{article} 
\usepackage[arxiv]{aaai23}  
\usepackage{times}  
\usepackage{helvet}  
\usepackage{courier}  
\usepackage[hyphens]{url}  
\usepackage{graphicx} 
\usepackage{natbib}  
\usepackage{caption} 
\usepackage{algorithm}
\usepackage{algorithmic}

\usepackage{newfloat}
\usepackage{listings}
\usepackage{adjustbox}
\usepackage{bm}

\usepackage[normalem]{ulem}
\usepackage{color}

\DeclareCaptionStyle{ruled}{labelfont=normalfont,labelsep=colon,strut=off} 
\floatstyle{ruled}
\newfloat{listing}{tb}{lst}{}
\floatname{listing}{Listing}
\title{MED-SE: Medical Entity Definition-based Sentence Embedding}
\author {
    Hyeonbin Hwang\textsuperscript{\rm 1},
    Haanju Yoo\textsuperscript{\rm 2},
    Yera Choi\textsuperscript{\rm 2,*}
}
\affiliations {
    \textsuperscript{\rm 1}Korea Advanced Institute of Science \& Technology (KAIST), South Korea\\
    \textsuperscript{\rm 2}NAVER CLOVA AI Lab, South Korea\\
    hbin0701@kaist.ac.kr, \{haanju.yoo, yera.choi\}@navercorp.com
}

\begin{document}

\maketitle

\begin{abstract}
We propose Medical Entity Definition-based Sentence Embedding (MED-SE), a novel unsupervised contrastive learning framework designed for clinical texts, which exploits the definitions of medical entities. To this end, we conduct an extensive analysis of multiple sentence embedding techniques in clinical semantic textual similarity (STS) settings. In the entity-centric setting that we have designed, MED-SE achieves significantly better performance, while the existing unsupervised methods including SimCSE show degraded performance. Our experiments elucidate the inherent discrepancies between the general- and clinical-domain texts, and suggest that entity-centric contrastive approaches may help bridge this gap and lead to a better representation of clinical sentences.
\end{abstract}

\begin{figure*}[t]
\centering
\includegraphics[width=\textwidth,height=\textheight,keepaspectratio]{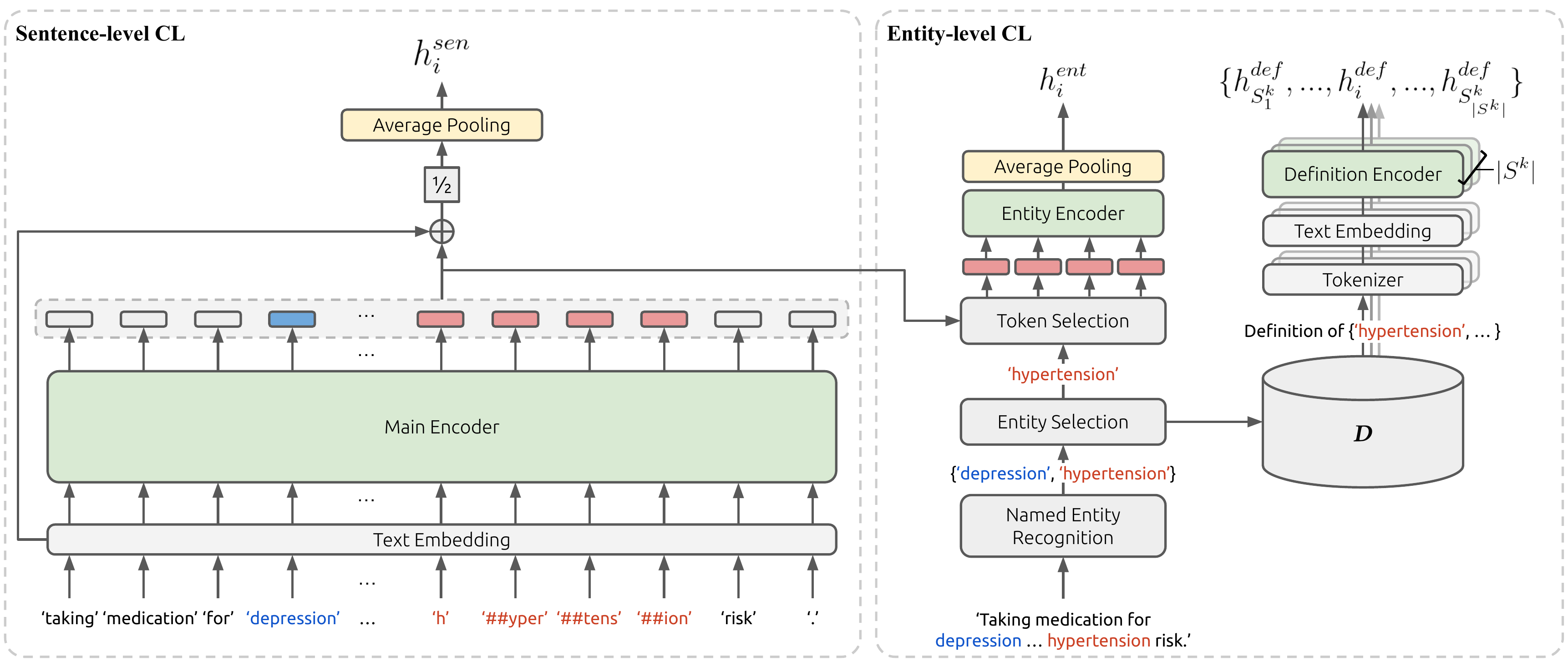}
\caption{Overview of MED-SE. We add an entity-level contrastive learning module to the original SimCSE framework, for which we extract entities of interest from the sentence embeddings and apply contrastive learning with its corresponding definitions embeddings.}
\label{fig1}
\end{figure*}

\section{Introduction}
Upon the success of large-scale pretrained language models (PLMs) for various tasks in natural language processing (NLP), many PLM-based studies focusing on the clinical domain have sought to facilitate the use and understanding of clinical notes, which contain unstructured yet rich medical information. Language models (LMs) pretrained on biomedical/clinical corpora are shown to outperform general-domain models in multiple aspects of clinical NLP. However, there exist unique and challenging issues for clinical texts, such as capturing the meaning of clinical sentences. Adequately representing clinical sentences is of particular practical interest, since it can contribute to alleviating clinicians' needs for condensed but accurate information from dispersed and heterogeneous clinical notes.

A high-quality clinical sentence embedding yet remains a challenging objective, as clinical notes are notorious for the heterogeneity of expressions to denote similar semantics \cite{Chen2019BioSentVecCS} and for the redundancy induced by copy-pasted text \cite{Weis2014CopyPA}. Unlike general-domain texts, clinical texts tend to be structurally formulaic and share common or similar text snippets due to frequent copying, pasting, and cloning \cite{Weis2014CopyPA}. Furthermore, the presence of medical jargons hinders thorough understanding without expert knowledge \cite{Mullenbach2021CLIPAD}. The situation becomes aggravated when medical jargons are the only source of difference between similarly constructed sentences, as those sentences should be distinguished from each other solely based on the semantic discrepancy between complex medical concepts. Nonetheless, existing literature has not placed much attention on the potential influence of these characteristics of clinical texts on different sentence embedding strategies.

Driven by these observations, we propose Medical Entity Definition-based Sentence Embedding (MED-SE), a novel unsupervised contrastive learning (CL) framework that exploits the formal definitions of medical entities, which induces an improved representation of clinical sentences. We argue that the use of definitions provides additional knowledge on medical concepts that can function as \textit{anchors} retaining relatively stable semantics across clinical sentences, as suggested by \citet{nishikawa-etal-2022-ease} where the hyperlinks in general-domain texts were used as anchors. Through experiments in an \textit{entity-centric} clinical semantic textual similarity (STS) setting that we have designed, we show that our framework contributes to an enhanced understanding of clinical sentences with \textit{and} without medical entities.

To investigate the effectiveness of our approach, we conduct an extensive comparative analysis of MED-SE and recent unsupervised CL models, namely CT-BERT \cite{Carlsson2021SemanticRW} and SimCSE \cite{gao-etal-2021-simcse}, which yield successful results in the general domain. We also examine post-processing techniques including whitening \cite{Huang2021WhiteningBERTAE} and normalizing flow \cite{Li2020OnTS}, along with other augmentation strategies \cite{Yan2021ConSERTAC}. Notably, MED-SE achieves the highest performance in the entity-centric clinical STS setting and demonstrates significant advantages for clinical sentence embedding.

In summary, our contributions are as follows:
\begin{itemize}
\item We present MED-SE, a novel unsupervised CL framework tailored to the representation of clinical sentences, which utilizes the definitions of medical entities.
\item We examine multiple sentence embedding strategies through extensive experiments and identify the ones particularly effective for clinical texts.
\item We show evidence that accounting for the characteristics of clinical texts provides substantial advantages for clinical sentence embedding.
\end{itemize}

\section{Related Works}

\subsection{Sentence Embeddings in the Biomedical Domain}
There exist a number of publicly available clinical sentence embeddings, namely BioSentVec \cite{Chen2019BioSentVecCS} and BlueBERT \cite{Peng2019TransferLI}, which are derived from models trained on large-scale biomedical corpora encompassing PubMed abstracts \cite{Fiorini2018HowUI} and/or MIMIC-III clinical notes \cite{Johnson2016MIMICIIIAF}. These models are outside the scope of the current study, as we focus on developing a sentence embedding framework that primarily leverages clinical sentences containing medical entities and the corresponding entity definitions.


\subsection{Post-Processing Techniques to Improve Sentence Embeddings}

There have been studies featuring post-processing techniques to improve PLM-derived sentence embeddings. This line of research emphasizes alignment and uniformity in the embedding space. To our knowledge, the effects of these techniques on clinical sentence embedding have not been much explored. 

\subsubsection{Normalizing Flows}

\citet{Li2020OnTS} report an important finding that the sentence embedding distribution derived from BERT \cite{Devlin2019BERTPO} tends to be non-smooth and anisotropic, which in turn results in poor STS performance. They also present BERT-flow, a method to overcome this issue by transforming the anisotropic distribution into an isotropic Gaussian distribution with normalizing flows.

\subsubsection{Whitening, Token Averaging, and Layer Combination}
\citet{Huang2021WhiteningBERTAE} propose WhiteningBERT, which implements vector normalization via whitening to ameliorate the anisotropy in sentence embeddings. Whitening is a linear transformation which normalizes a vector with a given covariance matrix by converting it into another vector with an identity covariance matrix. They also compare the combinations of strategies to obtain sentence embeddings from PLMs. According to their findings, applying whitening, averaging token representations, and combining the top and bottom layers are the effective strategies.

\subsection{Sentence Embedding Using Contrastive Learning (CL)}
CL-based models seek to obtain improved sentence embeddings, with the ingenious use of positive (i.e., semantically similar) and negative (i.e., semantically divergent) embedding pairs. In the current study, we put special interest on works featuring unsupervised/self-supervised approaches that do not require labeled sentences, as they are more applicable to clinical settings in which labeling is highly time- and resource-consuming.

\subsubsection{CT-BERT}
In CT-BERT \cite{Carlsson2021SemanticRW}, the proposed Contrastive Tension (CT) framework implements two independent BERT-based models that maximize the dot product between the representations of identical sentences and minimize that between the representations of different sentences.

\subsubsection{SimCSE}
Recently, SimCSE \cite{gao-etal-2021-simcse} and its variants including additional submodules, such as EASE \cite{nishikawa-etal-2022-ease}, have demonstrated promising results for unsupervised sentence embedding in the general domain. SimCSE constructs positive pairs by feeding a single sentence twice into the encoder with different dropout masks, and negative pairs by leveraging other in-batch sentences.

Despite its success in the general domain, SimCSE's performance for clinical texts can be limited in two aspects. First, non-identical clinical sentences may be semantically interconnected with each other, especially through shared medical entities. Thus, a naive sampling strategy inattentive to these entities cannot but lead to negative pairs of varying qualities and eventually to suboptimal sentence representations. Second, many clinical sentences are distinguished from each other primarily on medical entities due to copy-pasting practices. This further exacerbates the difficulty of inferring the meanings of medical jargons solely based on the context. As SimCSE primarily concentrates on enhancing sentence-level representations, it may not necessarily contribute to a better understanding of such entity-oriented clinical sentences.

\subsubsection{EASE}
Our work is inspired by EASE \cite{nishikawa-etal-2022-ease}, which performs contrastive learning with entities in addition to the original SimCSE training objective. EASE constructs pairs of a sentence and a semantically related entity (positive entity) $D = {(s_i, e_i)}^m_{i=1}$, by leveraging hyperlink entities of each sentence from Wikipedia. Their objective is designed with ends of predicting a learnable entity embedding $e_i$.

Our work primarily differs from EASE in that we perform \textit{entity-level CL}, while EASE conducts \textit{CL with entities}. In other words, we seek to cater token representations of entity (from sentence embedding) to corresponding external knowledge source, rather than predicting entity itself with the whole sentence representation. By doing so, we enforce an entity-level matching scheme, opting for a context-invariant knowledge injection. This framework will enable a more flexible positioning of anchors than using entity embeddings. We also expect that these anchors contribute to a better understanding of contextual associations between different clinical sentences.

\section{Methods}

As illustrated in Figure \ref{fig1}, our model MED-SE is composed of two distinct compartments: (1) sentence-level CL and (2) entity-level CL modules.
We describe the two modules in detail in the following subsections.

\subsection{Sentence-Level CL}
\label{subsec:sencl}
The sentence-level CL objective of our framework  directly adopts that of unsupervised SimCSE. Let $B^k$ denotes a set of indices of sentences in the $k$-th batch, the sentence-level CL objective of the $k$-th batch is defined by

\begin{equation}
\mathcal{L}^{k, sen} = - \sum_{i \in B^k} \textmd{log}\frac{e^{\textmd{sim}(h_i, h_i^+)} / \tau}{\sum_{j \in B^k} e^{\textmd{sim}(h_i, h_j^+)} / \tau},
\label{eqn:simCSE-eq}
\end{equation}
where $\tau$ is the temperature hyperparameter, \begin{math}\textmd{sim}(s_1, s_2)\end{math} is the cosine similarity \begin{math}\frac{s^\top_1 s_2}{\|s_1\|\cdot\|s_2\|}\end{math}, and $h^+_i$ indicates an augmented sentence embedding of $h_i$ generated by our main encoder applying a different dropout mask.

\subsection{Entity-Level CL}
\label{subsec:entcl}
For MED-SE, we design an entity-level CL framework that injects additional knowledge with definitions to enhance the encoder's comprehension of medical jargons. Utilizing definitions allows several advantages compared to other widely used methods such as knowledge base, as it is generally more accessible, easy to implement, and does not require any pretraining.  

Let $m_{D,i}$ be an arbitrary medical entity of our interest, and $s_{D,i}$ be the sentence defining $m_{D,i}$. Then, we retrieve the definition of medical entity in each sentence from a pre-built dictionary $D = \{ (m_{D,i}, s_{D,i}) | i = 1, 2, 3, ... \}$. Thus, given an input sentence containing $m_{D,i}$, we apply CL objective between entity embedding derived from the sentence embedding and the definition embedding $s_{D,i}$ of that entity. 

\begin{table}[]
\centering
\caption{Statistics on medical entities}
\label{tab:entstats}
\begin{tabular}{lr}
\hline
\multicolumn{2}{c}{\textbf{Entity Statistics}}           \\ \hline
\# of Unique Entities    & \multicolumn{1}{r}{440}              \\
\# of Top 10\% Entities          & 325,999 \\
Total \# of Entities       & \multicolumn{1}{r}{433,318}           \\\hline

Per-sentence \# of Entities & \multicolumn{1}{r}{$1.92\pm 1.54$} \\ \hline
\end{tabular}
\end{table}

\begin{table}[]
\centering
\caption {Distribution of the sentence sets derived from MIMIC-III Discharge Summaries: all sentences ($S^{all}$), those with ($S^{ent}$) and without entities ($S^{none}$).}
\begin{tabular}{lrr}
\hline
\textbf{Set Type}     & \textbf{\# of Sentences} & \textbf{Percentage} \\ \hline
$S^{all}$                    & 3,076,144                & 31.88\%             \\
$S^{ent}$                    & 328,338                  & 3.40\%              \\
$S^{none}$                    & 2,418,559                & 25.06\%             \\ \hline
\textbf{Discharge Summaries} & 9,649,630                & 100\%               \\ \hline
\end{tabular}
\label{sent_statistics}
\end{table}

\subsubsection{Pipeline}
We use \textbf{MedCAT} \cite{Kraljevic2019MedCATM}, an off-the-shelf named entity recognition (NER) model, against each $i$-th input sentence $s_i$, to obtain a list of medical entities $M_i = \{m_{i, 1}, m_{i, 2}, ... \}$ and their corresponding token indices. We specifically perform NER for specific semantic types, \textbf{Disease or Syndrome} (T047) and \textbf{Mental or Behavioral Dysfunction} (T048), in light of the observation that medical interests arise primarily around symptoms and problems \cite{askphysician}. We denote the set of sentences \textit{without} these entities as \textbf{$S^{none}$} $ = \{s_i \ | \forall i \textrm{ s.t. } |M_i| = 0\}$. Then, by searching elements of $M_i$ in $D$, we get $M^D_i = M_i \cap D^{ent}$ 
where $D^{ent}$ is the set of all entities in $D$.

Subsequently, we define a subset of sentences $S^{ent} = \{s_i | \forall i \textrm{ s.t. } 1 \leq | M^D_i | \}$, in which every sentence contains at least one entity of interest. In the $k$-th batch, we randomly select a single entity among $M_i$ for each 
$s_i \in S^k = \{s_i |\  \forall i \ \in B_k \textrm{ s.t. } |M_i| \geq 1\}$
, and retrieve the definition of each chosen entity from $D$. We feed the resulting set of definitions to a separate definition encoder which outputs $h^{def}$. 

Meanwhile, with the sentence embeddings for k-th batch $h^{sen,k}$ from the main encoder in \ref{subsec:sencl}, we select the embeddings corresponding to the sentences in $S^k$. Afterwards, we extract the entity tokens from these embeddings using each assigned entity's in-sentence token index (or indices), and feed them into an entity encoder followed by average pooling, to get $h^{ent}$. Finally, our entity-level CL is applied between $h^{def}$ and $h^{ent}$. For the $k$-th batch, this is defined as

\begin{equation}
\mathcal{L}^{k, ent} = - \sum_{i \in S^k} \textmd{log}\frac{e^{\textmd{sim}(h^{ent}_i, h^{def}_i))} / \tau}{\sum^{|S^k|}_{j=1} e^{\textmd{sim}(h^{ent}_i, h^{def}_j)} / \tau}.
\label{def-eq}
\end{equation}

The overall loss function for the $k$-th batch is a weighted sum of the two aforementioned losses.

\begin{equation}
\mathcal{L}^k = \mathcal{L}^{k,sen} + \lambda \cdot \mathcal{L}^{k,ent}.
\label{overall-loss}
\end{equation}

\subsubsection{Sentence Embedding Pooling}
As the preceding works implement different strategies to pool sentence representations, we perform a preliminary experiment to compare three well-known strategies: (1) using the [CLS] token, (2) averaging the output from the last layer, and (3) averaging the output from the first and last layers. Notably, despite the discrepancies in pretraining corpora, all baseline models show the best performance with (3). Based on this observation, we apply this pooling method for the main experiments.

\subsubsection{Entity Pooling}
Medical entities are often composed of multiple tokens, and our entity embedding is obtained based on the corresponding token indices from the sentence embedding. Accordingly, there is a high chance that the resulting entity embedding consists of multiple tokens. To convert such multi-token entity embedding into a single-token one, we utilize an entity encoder and a pooler. In the main experiments, we employ a shallow Transformer encoder for the entity encoder and average pooling for the pooler.

\subsubsection{Treatment of Duplicate Entities}

As shown in Table \ref{tab:entstats}, we observe a severe imbalance in the occurrence frequency of entities, as the top 10\% entities comprise more than 75\% of all occurrences. We empirically confirm that there is an over 99.96\% chance of having one or more duplicate entities in a batch, by conducting 1-million steps of training simulation on $S^{ent}$. We also find out that in average, only 78.4\% of entities in a batch are unique. On the contrary, CL-based clinical sentence embedding warrants particularly careful selection of in-batch negative pairs, since different sentences that share the same entity are more likely to be associated in meaning. Thus, for the main MED-SE models, we replace sentences containing duplicate entities with randomly retrieved sentences $s_i \in S^{ent}$ with different entities, to ensure all sentences within a batch have discrete entities assigned.

\section{Experiments}
\subsection{Training Dataset}
\subsubsection{MIMIC-III Discharge Summaries}
Medical Information Mart for Intensive Care III (MIMIC-III) \cite{Johnson2016MIMICIIIAF} is a dataset collected across critical care units at Boston-area hospitals from 2001-2012, consisting of electronic health record data from 53,423 patients.

For the current study, we utilize the MIMIC-III discharge summaries, which contain rich information acquired between admission and discharge.
We parse the discharge summaries using \textbf{ScispaCy} \cite{scispacy_Neumann_2019} to retrieve approximately 9 million sentences. Then, we filter out sentences that are deemed too short (i.e., with less than 10 words) and obtain a resulting set of about 3 million sentences. We denote this base set of sentences as $S^{all}$, which is divided into the sentences with ($S^{ent}$) or without ($S^{none}$) the entities of interest, as described in \ref{subsec:entcl}.

To investigate whether the model performances vary according to the entity-orientedness of training sentences, we conduct experiments with different pretraining sets: (1) $S^{all}$ for the general setting, (2) $S^{ent}$ for the entirely entity-centric setting, and (3) mixtures of $S^{ent}$ and $S^{none}$. We describe the distribution of the sentence sets in Table \ref{sent_statistics}.

\subsection{Evaluation Dataset}
\subsubsection{MedSTS/ClinicalSTS}
MedSTS \cite{Wang2020MedSTSAR} consists of different versions of datasets created for clinical STS. We use the version released as a part of the 2019 n2c2/OHNLP Track on Clinical Semantic Textual Similarity (\textbf{ClinicalSTS 2019}; \citet{Wang2020The2N}), which consists of 1,642 training pairs and 412 test pairs collected from Mayo Clinic’s clinical data warehouse. The semantic similarity of each sentence pair was rated with a score of 0-5 by two medical experts.

For each pair, we compute the cosine similarity of the output embeddings and directly report the Spearman's rank order correlation coefficient (SROCC) with the human-annotated scores. The SROCC is computed with the following formula
\begin{equation}
\rho = 1- {\frac {6 \sum d_i^2}{m(m^2 - 1)}},
\end{equation}
where
$d$ is the distances of the ranks between pairwise sentences,
and $m$ is the total number of samples.

\subsection{Models}
For comparison with our model, we adopt the following models: (1) baseline models including BERT \cite{Devlin2019BERTPO}, BioBERT \cite{Lee2020BioBERTAP}, and ClinicalBERT \cite{Alsentzer2019PubliclyAC}, and (2) recent unsupervised CL models including CT-BERT \cite{Carlsson2021SemanticRW} and SimCSE \cite{gao-etal-2021-simcse}. We also test the effects of post-processing methods including whitening \cite{Huang2021WhiteningBERTAE} and normalizing flows \cite{Li2020OnTS}. In addition, we apply a set of augmentation strategies from ConSERT \cite{Yan2021ConSERTAC} including token cutoff and shuffling, to replace the dropout strategy of SimCSE.

\begin{table*}[tt]
\centering
\caption{General overview of model performances on the MedSTS Train and Test sets with different training sets: all sentences ($S^{all}$), sentences with entities ($S^{ent}$), and a 9:1 mixture of $S^{ent}$ and $S^{none}$ ($S^{0.9ent}$), respectively. Every MED-SE model adopts a Transformer encoder for word pooling, except for the one using a cross-attention module (CA). All LM baselines are directly evaluated on the MedSTS Train and Test sets. Models indicated with $\dagger$ use the [CLS] Token, while those denoted with $^*$ use the first-last layer average for sentence embedding pooling.}
\begin{adjustbox}{width=0.95\textwidth}
\begin{tabular}{l|rr|rr}
\hline
\rule{0pt}{2ex} \def\arraystretch{1.4}%
\textbf{Models}                                & \multicolumn{1}{r}{\textbf{MedSTS Train Set}} &&  \multicolumn{1}{r}{\textbf{MedSTS Test Set}}  \\ \hline
\textbf{Language Model Baselines}     & & & &                                                                                                                        \\ \hline
BERT*                                & & 0.648 &                        & 0.392                                  \\
BioBERT*                              & & 0.624                                  & & 0.445                               \\
ClinicalBERT*                    & &  0.638 &                               &   0.450                               \\
+ Whitening                                          &      & 0.563                                & & 0.456\\ \hline
                                  \textbf{Unsupervised CL Models}                         & S\textsubscript{all}                        & S\textsubscript{ent}                         & S\textsubscript{all}                        & S\textsubscript{ent}                        \\ \hline
    CT-BERT \textdagger                                                   &  $0.705_{\pm0.012}$                           &  $0.657_{\pm0.005}$                      &    $0.665_{\pm0.036}$                         & $0.513_{\pm0.013}$ \\
    CT-BERT *                                                    &  $0.723_{\pm0.009}$                           &   $0.660_{\pm0.011}$                    &           $0.711_{\pm0.008}$                  &    $0.528_{\pm0.025}$                         \\
SimCSE \textdagger                                         & $0.716_{\pm0.016}$          & $0.560_{\pm0.017}$           & $0.715_{\pm0.014}$             & $0.450_{\pm0.027}$          \\
SimCSE *                              & $0.742_{\pm0.001}$ & $0.673_{\pm0.008}$           & $0.716_{\pm0.006}$             & $0.571_{\pm0.028}$          \\
Token Cutoff (from ConSERT) *                             & $0.742_{\pm0.002}$          & $0.671_{\pm0.010}$       & $\bm{0.735}_{\pm0.017}$   & $0.552_{\pm0.038}$          \\
Token Shuffle (from ConSERT) *                           & $\textbf{0.743}_{\pm0.024}$          & $0.671_{\pm0.009}$           & $0.706_{\pm0.007}$               & $0.565_{\pm0.026}$          \\
MED-SE \textdagger                                        & $0.689_{\pm0.003}$          & $0.624_{\pm0.021}$           & $0.702_{\pm0.012}$          & $0.533_{\pm0.031}$          \\
MED-SE *                               & $0.735_{\pm0.004}$          & $0.702_{\pm0.005}$           & $0.732_{\pm0.014}$          & $0.627_{\pm0.020}$         \\
MED-SE * + Replace Duplicate Ent.      & $0.736_{\pm0.003}$          & $0.721_{\pm0.008}$           & $0.728_{\pm0.006}$          & $0.687_{\pm0.035}$          \\
MED-SE * + Replace Duplicate Ent. (CA) & $0.737_{\pm0.003}$          & $\bm{0.728}_{\pm0.006}$ & $0.717_{\pm0.013}$          & $\bm{0.735}_{\pm0.018}$ \\ \hline
\textbf{Best Setting for $S^{ent}-S^{none}$ Ratio}           & & S\textsubscript{0.9ent}                                 & \multicolumn{2}{r}{S\textsubscript{0.9ent}}                                 \\ \hline
SimCSE * + Replace Duplicate Ent. + 90\% $S^{ent}$              & & $\bm{0.745}_{\pm0.003}$            & \multicolumn{2}{r}{$0.730_{\pm0.004}$}                    \\
MED-SE * + Replace Duplicate Ent. + 90\% $S^{ent}$             &  & $0.744_{\pm0.003}$                     &  & $\bm{0.767}_{\pm0.005}$ \\ \hline
\end{tabular}
\end{adjustbox}
\label{table_overview}
\end{table*}

\begin{table}[]
\centering
\rule{0pt}{2ex} \def\arraystretch{1.2}%
\caption{MED-SE vs. SimCSE performance with varying ratio of $S^{ent}$ and $S^{none}$ in the training corpus. All sentences with duplicate entities are replaced. \\}
\begin{adjustbox}{width=0.95\textwidth / 2}
\begin{tabular}{l|ll|ll}
\hline
     & \multicolumn{2}{c|}{\textbf{MedSTS Train}} & \multicolumn{2}{c}{\textbf{MedSTS Test}} \\ \hline
$S^{none} \%$     & MED-SE          & SimCSE          & MED-SE         & SimCSE         \\ \hline
0\%   &   $0.721_{\pm0.008}$              &   $0.685_{\pm0.007}$              & $0.687_{\pm0.035}$               &   $0.560_{\pm0.017}$            \\
3\%   &  $0.726_{\pm0.004}$               &    $0.706_{\pm0.004}$             &  $0.694_{\pm0.012}$              &   $0.662_{\pm0.020}$             \\
6\%   &  $0.741_{\pm0.056}$               &   $0.730_{\pm0.006}$              & $0.758_{\pm0.016}$               &   $0.713_{\pm0.009}$            \\
10\%  &     $\bm{0.744}_{\pm0.003}$            &    $\bm{0.745}_{\pm0.003}$             &  $\bm{0.767}_{\pm0.005}$              &      $\bm{0.730}_{\pm0.004}$          \\
25\%  &  $0.745_{\pm0.003}$               & $0.746_{\pm0.003}$                &    $0.750_{\pm0.006}$            &  $0.727_{\pm0.012}$              \\
50\%  &  $0.738_{\pm0.003}$               & $0.747_{\pm0.003}$                 &  $0.732_{\pm0.015}$              &    $0.724_{\pm0.014}$            \\
75\%  & $0.736_{\pm0.002}$        &   $0.744_{\pm0.003}$         &    $0.732_{\pm0.011}$           & $0.726_{\pm0.016}$                \\
100\% &   -            &    $0.738_{\pm0.004}$             &   -            &   $0.716_{\pm0.011}$             \\ \hline
\end{tabular}
\end{adjustbox}
\label{tab:ratio}
\end{table}

\begin{table}[]
\centering
\rule{0pt}{2ex} \def\arraystretch{1.5}%
\caption{Results of perturbations on the entity-level CL mechanism of MED-SE. \\} 
\begin{adjustbox}{width=0.95\textwidth / 2}\begin{tabular}{l|ll|ll}
\hline
                      & \multicolumn{2}{l|}{\textbf{MedSTS Train}} & \multicolumn{2}{l}{\textbf{MedSTS Test}}   \\ \hline
\textbf{Methods}               & Naive           & Replace    & Naive           & Replace   \\ \hline
Random Tokens         & $0.702_{\pm0.003}$          & $0.702_{\pm0.003}$          & $0.653_{\pm0.009}$          & $0.667_{\pm0.006}$          \\
Shuffle Sentences         & $0.668_{\pm0.006}$          & $0.679_{\pm0.019}$          & $0.535_{\pm0.030}$          & $0.560_{\pm0.049}$         \\
Sentence-Pooling      & $\bm{0.712}_{\pm0.007}$ & $0.706_{\pm0.003}$          & $\bm{0.671}_{\pm0.018}$         & $0.682_{\pm0.014}$ \\ \hline
Entity-Pooling & $0.701_{\pm0.054}$          & $\bm{0.721}_{\pm0.008}$ & $0.627_{\pm0.196}$ & $\bm{0.687}_{\pm0.035}$ \\ \hline
\end{tabular}
\end{adjustbox}
\label{perturb}
\end{table}

\subsection{Setup}
We assess the model performance by evaluating the embedding similarity scores on the MedSTS Train and Test sets for every $n$-th step ($n$=25). Because we observe that the respective performances for the MedSTS Train and Test sets tend to diverge in the baseline models, we select the step $\hat{s}$ for which the SROCC for the MedSTS Train set ($\rho^{\hat{s}}_{train}$) is the highest, and report both $\rho^{\hat{s}}_{train}$ and $\rho^{\hat{s}}_{test}$. 

For MED-SE, we choose a batch size of 32 and maximum token lengths of 128, which yields the best outcome for the MedSTS Train set. We set the weight of the entity-level CL loss $\lambda$ to 0.1. Also, we adopt a Transformer encoder as the entity encoder and the first-last layer average for sentence embedding pooling. Separate BioBERT v1.0 models are implemented as the main and definition encoders. For the predefined dictionary, we use the Experimental Factor Ontology database (Malone et al. 2010) from the Ontology Lookup Service of European Bioinformatics Institute (Jupp et al. 2015).

For all other models, we adopt the hyperparameter setting presented in their respective work. We report the mean and the standard deviation of the results from five runs with different random seeds. Although we train all models for 2,000 steps, in most cases, the models quickly reach their respective best SROCC for the MedSTS Train set and then overfit to MIMIC-III. All our experiments are carried out in a single GPU setting, using either A100 80 GB or A100 40 GB.

\section{Results}

\subsection{Main Results}

In Table \ref{table_overview}, we report the comparative performance of the baseline LM models, unsupervised CL models, and MED-SE. As expected, MED-SE outperforms other models by far when $S^{ent}$ is used as the training corpus. In the case of $S^{all}$, it suffers a slight disadvantage compared to other unsupervised CL models, which is likely due to the instability caused by the varying size of $S^k$ (which may even reduce to 0) in different iterations. Notably, for the setting in which $S^{ent}$ and $S^{none}$ are mixed by a fixed ratio of 9:1, our method shows the highest performance among all models on MedSTS Test set.

In addition, the token cutoff strategy scores the highest in the MedSTS Test set when trained with $S^{all}$, which contradicts the main observation from SimCSE. Therefore, we call for a further exploration of augmentation strategies on clincial texts, in order to disclose the distinctive characteristics of clinical texts more precisely.

\subsubsection{Adjusting the Ratio of Sentences without Entities ($S^{none}$)}

Table \ref{tab:ratio} depicts an interesting finding that a small injection of $S^{none}$ leads to drastic changes in performance, especially when going from 3\% to 6\%. The proportion of 10\% appears to be the best empirical threshold, which allows for the models to learn the most encompassing representations by facilitating $S^{ent}$ to better leverage the relatively \textit{harder} negatives from $S^{none}$. 

When we train only with $S^{ent}$, the performances are generally much lower, which confirms the model's difficulty in understanding these jargons. In contrast, a rigorous understanding of these words is essential in representing the overall sentence in an entity-aware manner. Our method shows particular strength at this point, as MED-SE clearly outperforms SimCSE in all settings where $S^{ent}$ is higher, with a conspicuous superiority at the setting of 0\%. ($\textbf{+0.034}$, $\textbf{+0.125}$ improvements for the Train and Test sets, respectively). At the same time, as training solely with $S^{none}$ shows a weaker performance than training with ratio of 3\% to 75\%, we cannot emphasize enough the importance of $S^{ent}$ and its entities when learning diverse representations from medical domain.

\begin{table}[]
\centering
\caption{Results with varying weights of the entity-level CL task, when trained on $S^{ent}$.}
\begin{adjustbox}{width=0.95\textwidth / 2}
\rule{0pt}{2ex} \def\arraystretch{1.3}%
\begin{tabular}{llll}
\hline
\textbf{Entity-level CL $\lambda$}   & \multicolumn{1}{r}{$0.001$} & \multicolumn{1}{r}{$0.01$} & \multicolumn{1}{r}{$0.05$} \\ \hline
\textbf{MedSTS Train}            & $0.708_{\pm0.007}$          & $0.715_{\pm0.006}$          & $0.716_{\pm0.008}$ \\
\textbf{MedSTS Test}              & $0.628_{\pm0.025}$         & $0.650_{\pm0.018}$         & $0.668_{\pm0.022}$  \\ \hline
\multicolumn{1}{r}{$0.1$}  & \multicolumn{1}{r}{$0.2$}  & \multicolumn{1}{r}{$0.25$} & \multicolumn{1}{r}{$0.5$}  \\ \hline
\multicolumn{1}{r}{$0.721_{\pm0.008}$} & $0.722_{\pm0.005}$ & $0.723_{\pm0.004}$ & $0.717_{\pm0.003}$          \\
\multicolumn{1}{r}{$0.687_{\pm0.035}$} & $0.688_{\pm0.008}$ & $0.691_{\pm0.007}$ & $0.675_{\pm0.015}$          \\ \hline
\end{tabular}
\end{adjustbox}
\label{def_weight}\end{table}

\subsubsection{Verifying Entity-Level CL Mechanism}

To verify whether our model can take advantage of the definitions through entity-level CL objective, we inject perturbations by randomly assigning token indices for the entities, or try shuffling sentence embeddings for $S^k$ so that resulting embedding in $h^{ent}$, and $h^{def}$ do not match each other. We also attempt pooling the entire sentence representation instead of indexing out the entity tokens, and report the results in Table \ref{perturb}. We see that matching entity definitions with the wrong entities drastically lowers the performances, supporting that definitions are constructive sources for our cooperative CL framework. On the other hand, word-level random indexing and pooling sentence yield better results on Naive setting, which may mislead to a conclusion that matching scheme may not be necessarily done in entity level. However, it is worth to note that these methods outperform only in naive setting as they better elude the noise introduced by the sentences with duplicate entities. This is further evidenced by the minimal to no performance gain by these two approaches with proper replacement setting. Then, we finally observe our entity-pooling method outperforms all variants in both MedSTS train and test set.

\begin{table}[]
\centering
\vspace{-1.9em}
\rule{0pt}{2ex} \def\arraystretch{1.3}%
\caption{Comparison of strategies to handle in-batch duplicate entities when trained on $S^{ent}$. \\}
\begin{adjustbox}{width=0.95\textwidth / 2}
\begin{tabular}{l|ll|ll}
\hline
     & \multicolumn{2}{c|}{\textbf{MedSTS Train}} & \multicolumn{2}{c}{\textbf{MedSTS Test}} \\ \hline
\textbf{Strategies}      & MED-SE          & SimCSE          & MED-SE         & SimCSE         \\ \hline
Naive   &   $0.702_{\pm0.005}$              &   $0.661_{\pm0.006}$              & $0.627_{\pm0.020}$               &   $0.528_{\pm0.018}$            \\
Remove   &  $0.717_{\pm0.006}$               &    $0.676_{\pm0.014}$             &  $0.670_{\pm0.010}$              &   $0.554_{\pm0.020}$             \\
Replace  &  $\bm{0.721}_{\pm0.008}$               &   $\bm{0.685}_{\pm0.007}$             & $\bm{0.687}_{\pm0.04}$               &   $\bm{0.560}_{\pm0.017}$            \\
 \hline
\end{tabular}
\end{adjustbox}
\label{tab:entdup}
\end{table}

\begin{table}[]
\centering
\caption{Comparison of different encoding methods for entity embedding}
\rule{0pt}{2ex} \def\arraystretch{1.1}%
\begin{adjustbox}{width=0.8\columnwidth}
\begin{tabular}{l|l|l}
\hline
                 & \textbf{MedSTS Train}    & \textbf{MedSTS Test}  \\ \hline
Mean             & $0.721_{\pm0.007}$                  & $0.682_{\pm0.024}$                \\
Bi-LSTM          & $0.718_{\pm0.006}$                   & $0.673_{\pm0.023}$                 \\
FC Layer         & $0.722_{\pm0.006}$                  & $0.672_{\pm0.028}$              \\
Transformer Enc. & $0.721_{\pm0.008}$                   & $0.687_{\pm0.035}$                \\
Cross Attention  & $\bm{0.728}_{\pm0.007}$            & $\bm{0.734}_{\pm0.020}$         \\ \hline
\end{tabular}
\end{adjustbox}
\label{entpool}
\end{table}

\subsection{Ablation Studies}

\subsubsection{Adjusting the Weight on the Entity-Level CL}

We report the results from adjusting the weight on the entity-level CL loss in Table \ref{def_weight}. As the weight approaches 0, the performance gradually decreases, which once again shows the effectiveness of entity-level CL in presence of medical jargons. Also, we have attempted to incorporate MLM objective as in SimCSE; however, it resulted in performance degradation both with and without entity-level CL, which again suggests a need for a distinct approach to tackle clinical texts.

\subsubsection{Entity Encoder}
Although we adopt Transformer Encoder as Entity Encoder, we also test with other methods so that we can effectively represent multi-token medical entities from $h_{sen}$. We conduct experiments with four conventional variants including ours: (1) averaging corresponding tokens (Mean), (2) bidirectional LSTM, (3) fully connected layer, and (4) transformer encoder. We also test an extra cross-attention (CA) module proposed in \cite{CAT}, which asymmetrically combines two separate embedding sequences and computes the attention output of one embedding in the context of the other. As shown in Table \ref{entpool}, the CA module leads to the best result with $S^{ent}$. We speculate that it maximizes the effect of the context-invariant representation of medical entities, as we encode the entity tokens in the attentive context of the corresponding sentence and consequently coerce this consistency across different sentences. Nevertheless, in consideration of its limited performance with $S^{all}$ (Table \ref{table_overview}) and longer convergence time, we suggest that a Transformer encoder is a more accessible option in general.

\subsubsection {Duplicate Entities in contrastive learning}
We test our hypothesis on the adverse effects of entity duplicates by contrasting the following strategies: (1) training naively on $S^{ent}$ with entity duplicates, (2) removing the sentences containing duplicates, and (3) replacing them with other sentences containing disparate entities. As shown in Table \ref{tab:entdup}, replacing yields the best outcomes, with point increases of \textbf{0.024} and \textbf{0.032} with SimCSE and \textbf{0.019} and \textbf{0.060} with MED-SE, respectively for the MedSTS Train and Test sets. Although removing can be considered as an alternative, it may incur undesirable effects from the instability due to varying batch sizes.

\begin{table}[]
\centering
\rule{0pt}{2ex} \def\arraystretch{1.2}%
\caption{Results of MED-SE trained from various baselines, with $S^{ent}$ and replacement of duplicate entities \\}
\begin{adjustbox}{width=0.95\columnwidth}
\begin{tabular}{l|l|l|l}
\hline
                      & \textbf{Methods}   & \textbf{MedSTS Train} & \textbf{MedSTS Test} \\ \hline
\textbf{BERT}         & Baseline & 0.648                & 0.392               \\
                      & SimCSE   & $0.702_{\pm0.014}$               & $0.604_{\pm0.031}$              \\
                      & MED-SE   & $\textbf{0.703}_{\pm0.006}$                &  $\textbf{0.628}_{\pm0.015}$              \\ \hline
\textbf{BioBERT}     & Baseline & 0.624                & 0.445              \\
                      & SimCSE   & $0.685_{\pm0.007}$               & $0.560_{\pm0.017}$               \\
                      & MED-SE   & $\textbf{0.721}_{\pm0.008}$                & $\textbf{0.687}_{\pm0.035}$               \\ \hline
\textbf{ClinicalBERT} & Baseline & 0.638                & 0.450               \\
                      & SimCSE   & $0.702_{\pm0.006}$                & $0.624_{\pm0.015}$               \\
                      & MED-SE   & $\textbf{0.715}_{\pm0.007}$               & $\textbf{0.640}_{\pm0.019}$               \\ \hline
\end{tabular}
\end{adjustbox}
\label{baselines}
\end{table}

\section{Analysis}


\subsubsection{Different Baselines}
While we use BioBERT for the sentence encoder, many recent studies adopt ClinicalBERT, which further pretrained BioBERT on MIMIC-III. We conduct additional experiments to examine the effects of different baselines (Table \ref{baselines}). Although the baseline results attest to the superiority of ClinicalBERT over other models, a different tendency is observed when the models are pretrained on $S^{ent}$ with SimCSE or MED-SE CL objectives. MED-SE induce consistent performance gains over SimCSE across all baselines.
Interestingly, the performance improvement induced by MED-SE appears to be smaller for ClinicalBERT than for BioBERT. We suspect that, as ClinicalBERT has already been extensively saturated with MIMIC-III, it has become less sensitive to entity-centric learning objectives.

\subsubsection{Sentence Embedding Pooling}
Intriguingly, compared to the use of the [CLS] token, the first-last average pooling strategy has consistently yielded better results in all settings. As suggested by works on PLM layers \cite{Tenney2019BERTRT, Jawahar2019WhatDB} and thoroughly investigated by \citet{Huang2021WhiteningBERTAE}, layer combination can improve sentence embeddings by effectively exploiting the information hierarchy captured by different layers. Our results indicate that layer combination is a particularly effective strategy for clinical sentences, which are structurally formulaic and entity-oriented. We postulate that utilizing the information from the first and last layers allows to better reflect these unique traits of clinical sentences, possibly through the aggregation of the syntax- and semantics-oriented information respectively from the first and last layers, as suggested by \citet{Tenney2019BERTRT}.

\subsubsection{Alignment and Uniformity}
\begin{figure}
\centering
\includegraphics[width=0.95\columnwidth]
{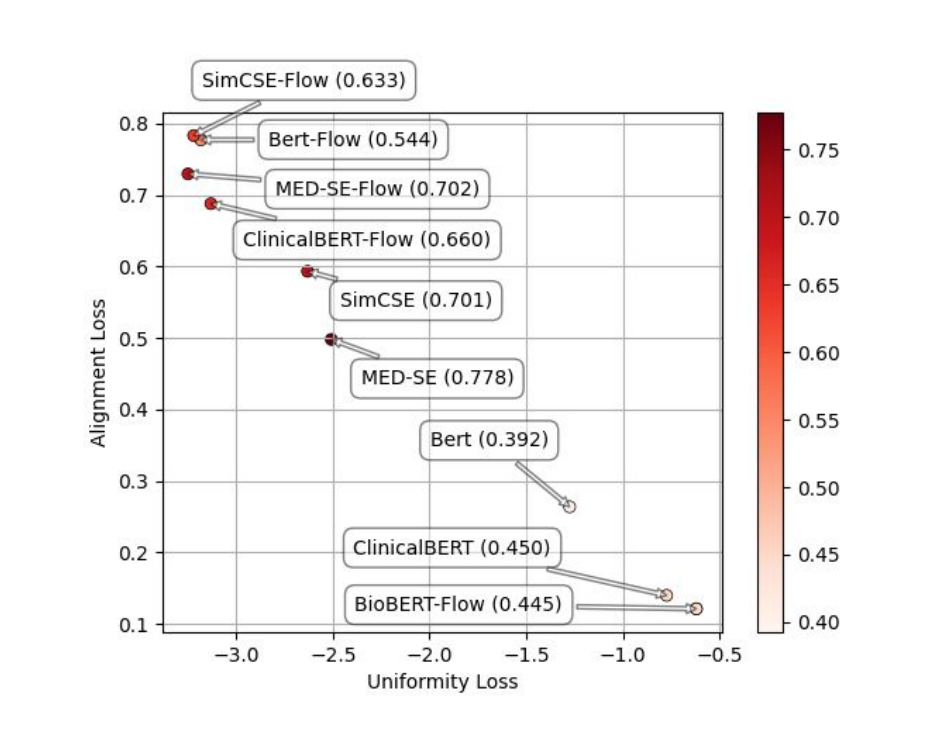}
\caption{Alignment and uniformity of different models. For both alignment and uniformity, lower values suggest more desirable outcomes. The values in parentheses indicate the Spearman's rank order correlation coefficients on the MedSTS Test set. The displayed SimCSE model is naively trained on $S^{all}$, while MED-SE is trained on $0.9S^{ent}$ with replacement of duplicate entities.}
\label{fig2}
\end{figure}

We also examine the performances of MED-SE and other models in terms of alignment and uniformity (Figure \ref{fig2}), which have been suggested as proxies of the sentence embedding quality in multiple preceding works \cite{Li2020OnTS, Huang2021WhiteningBERTAE, gao-etal-2021-simcse}. The outcomes appear to further emphasize the distinct advantages of MED-SE. With the combined use of entity-oriented sentences and the replacement strategy for duplicate entities, MED-SE achieves a good balance between alignment and uniformity, which is attested by the highest SROCC. We believe that this desirable balance between alignment and uniformity is grounded on the model's efforts to secure more relevant negative pairs in the context of medical entity semantics. Meanwhile, the results of other models, especially those implementing normalizing flows, strongly implicate that the alignment-uniformity imbalance is associated with degraded performance in clinical sentence embedding.

\section{Conclusion}
We present MED-SE, a novel unsupervised CL framework for clinical sentence embedding. In addition to the sentence-level CL objective adopted from SimCSE, MED-SE introduces an entity-level CL objective between the representations of within-sentence medical entities and those of entity definitions. With the entity-level CL module, MED-SE seeks to exploit the characteristics of clinical sentences and demonstrates clear advantages in clinical sentence embedding, especially when trained on entity-oriented sentences.

\appendix


\bibliography{main}

\end{document}